\documentclass[conference]{IEEEtran}
\usepackage[utf8]{inputenc}
\usepackage[english]{babel}
\usepackage[backend=bibtex,style=numeric]{biblatex}  
\usepackage{csquotes}
\usepackage{amsmath}
\usepackage{graphicx}
\usepackage{graphics}
\usepackage{moreverb}
\usepackage{diagbox}
\usepackage[colorlinks=true, citecolor=blue,linkcolor=blue,bookmarks=false]{hyperref}
\usepackage{epsfig}
\usepackage{hangcaption}
\usepackage{txfonts}

\usepackage{newclude}
\usepackage{float}
\usepackage{multirow}  
\usepackage[export]{adjustbox}
 \usepackage[labelsep=quad,indention=10pt]{subfig}
\PassOptionsToPackage{subfigure}{tocloft}
\usepackage{fixmetodonotes}

\usepackage{todonotes}

\makeatletter
\def\url@leostyle{%
  \@ifundefined{selectfont}{
    \def\UrlFont{\sf}
  }{
    \def\UrlFont{\small\bf\ttfamily}
  }}
\makeatother

\def\BibTeX{{\rm B\kern-.05em{\sc i\kern-.025em b}\kern-.08em
    T\kern-.1667em\lower.7ex\hbox{E}\kern-.125emX}}
\begin{document}

\title{Towards Evaluating Driver Fatigue with Robust Deep
Learning Models \\
}

\author{
\IEEEauthorblockN{Ken Alparslan\textsection}
\IEEEauthorblockA{\textit{Department of Computer Science} \\
\textit{Conestoga College}\\
Waterloo, ON, CA \\
kalparslan6724@conestogac.on.ca}
\and
\IEEEauthorblockN{Yigit Alparslan\textsection}
\IEEEauthorblockA{\textit{Department of Computer Science} \\
\textit{Drexel University}\\
Philadelphia, PA, US \\
ya332@drexel.edu}
\and
\IEEEauthorblockN{Dr. Matthew Burlick}
\IEEEauthorblockA{\textit{Department of Computer Science} \\
\textit{Drexel University}\\
Philadelphia, PA, US \\
mjb528@drexel.edu}
}
\maketitle
\begingroup\renewcommand\thefootnote{\textsection}
\footnotetext{These co-first authors contributed equally.}
\endgroup

\begin{abstract}
In this paper, we explore different deep learning based approaches to detect driver fatigue. Drowsy driving results in approximately 72,000 crashes and 44,000 injuries every year in the US \cite{s} and detecting drowsiness and alerting the driver can save many lives. There have been many approaches to detect fatigue, of which eye closedness detection is one. We propose a framework to detect eye closedness in a
captured camera frame as a gateway for detecting drowsiness. We explore two different datasets to detect eye closedness. We develop an eye model by using new Eye-blink dataset \cite{f14} and a face model by using the Closed Eyes in the Wild (CEW) \cite{f14}. We also explore different techniques to make the models more robust by adding noise. We achieve 95.84\% accuracy on our eye model and 80.01\% accuracy on our face model. We also see that we can improve our accuracy on the face model by 6\% via adversarial training and data augmentation. We hope that our work will be useful to the field of driver fatigue detection to avoid potential vehicle accidents related to drowsy driving.  

\end{abstract}


\section{Introduction}
Drowsy driving results in approximately 72,000 crashes and 44,000 injuries every year in the US \cite{s}. Driving after going more than 20 hours without sleep is the equivalent of driving with a blood-alcohol concentration of 0.08\% – the U.S. legal limit.
A driver is three times more likely to be in a car crash if fatigued compared to a non-fatigued driver \cite{wcpcd13}.The problem of drowsy driving is especially important for night-drivers and long-haul drivers. Some car manufacturers have started to implement pressure sensors on the steering wheel to detect whether the driver is engaged with the wheel or not, but they remained limited and such sensors are usually sold as luxury in a high-end sports car. We aim to develop a more robust, more cost effective and ways of detecting driver fatigue for everyone who drives, but not just for luxurious car owners. 

Machine learning can be used to solve hard computer vision problems such as crime prevention, surveillance, security \cite{aplkc20}. Driver fatigue detection is a challenging task, yet it includes many parts where machine learning can be applied to.  Learning when and how the driver is showing signs of fatigue is a task that heavily depends on face detection. Even though face detection and classification is a hard task, there have been several breakthroughs in the literature \cite{aab20audio} \cite{aakk20}. In this study, we explore building a deep learning model, one of many recent breakthroughs in the literature, in order to solve this challenging problem of facial landmark detection and then finally driver fatigue detection. While our main goal is to build a robust deep learning model with less false positives and negatives, we also consider the size aspect of the models. We also aim to develop a model that can potentially be used in a very constrained environment such as a mobile application because we foresee that a mobile application can be used by everyone for fatigue detection to alert drowsy drivers. Luxurious high-end car companies are implementing driver fatigue warning systems \cite{mercedesbenz} and providing built-in features to detect driver distractions, but not everyone can afford those cars due to their high price. By developing a model that can run in a mobile application, we are also making this alert system available to everyone with a mobile phone. A robust model is needed to reduce the false negatives as well as the false positives. False negatives are the cases where the driver is sleeping when driving but the model doesn't alert the driver. False positives are the cases where the model is alerting the driver even though the driver is alert and focused. Fatigue detection for drivers problem becomes non-trivial because peoples lives are involved and the model should be robust enough to eliminate false negatives as well as false positives. We also need to consider the fact that alerting drivers when not needed is as dangerous as failing to alert when needed. Alerting drivers when not needed can create unnecessary distraction that otherwise would have not existed. Avoiding false negatives and false positives requires us to build robust deep learning models for this task. Even though we are not directly aiming to create a mobile application, we should take into consideration the limited processing power and memory available at a mobile phone. Even though the Moore's empirical studies show that the memory and CPU get more cost effective each year, we still should bear in mind that our robust model should be able to fit into a mobile phone and run as a mobile application.

The task of eye closedness detection is to decide whether the eyes are closed \cite{wcpcd13}. Detecting eyes' closedness has a variety of applications. Driver fatigue detection is one of them. This task is a challenging problem since the degree of eye closedness may be different for each face and there are many ambient factors that may significantly change the appearance of the eyes, such as lighting, pose, scales and imaging conditions \cite{a96}
In addition, inaccurate eye localization may introduce a great difficulty to this problem. A recent push in the literature is to use deep learning based models in order to detect eyes closedness. One problem with the mentioned approach is that deep learning models can be also easily fooled as demonstrated by \cite{cw16} \cite{cw17} \cite{cw17bypassingtendetectionmethods} \cite{aajfacialadversarial}. Our goal in this paper is to create a robust fatigue detection framework with high accuracy. In this study, we also consider those perturbations such as bad lightning, pose, and shadows as adversarial attacks and try to provide solutions for the problem of robust eye closedness detection in real life via adversarial training.

 The task of eye localization is made more difficult by occlusions, problematic lighting, pose, and motion blur which can cause an adversarial setting for the model. Adversarial attacks are defined as perturbations to the sample image and they can be added on purpose in order to result in misclassification \cite{a19} \cite{a19transferableadversarials} under a white box setting \cite{pmwjs15distillation} or they can be added as a noise under black box setting \cite{m17} where the noise is not added on purpose but added because of the environment. For a machine learning model inside a Tesla 3 car model, a sticker on a stop sign can be defined as an black-box adversarial attack. In literature, those perturbations are considered black-box attacks \cite{cw16} \cite{cw17}. So, when we are training our models to detect eye closedness, we employ defenses in order to mitigate those attacks. Current literature has been focusing on adversarial training \cite{bp} \cite{kwt87} as a defense mechanism. We explore data augmentation methods and adversarial training in order to make our model more generalizable than what it would have been otherwise. One method that we use adversarial training \cite{ych92} \cite{ly96}. We define eight aspects on a driving setting that can be considered as a adversarial attack.

Adversarial attacks in the form of data augmentation during the training phase of the fatigue detection model can be summarized here:\\
\begin{enumerate}
    \item \textbf{Rotation:} Image can be rotated randomly depending on the driver's position and the camera angle
    \item \textbf{Width Shift:} Image width might depend on the camera angle. The model needs to mitigate this randomness 
    \item \textbf{Height Shift:} Image height might depend on the camera angle. The model needs to mitigate this randomness
    \item \textbf{Shear angle Shift:} Drivers plane intersects with the plane in which camera is mounted on a car. This creates additional randomness and the model needs to mitigate this randomness
    \item \textbf{Zoom:} The camera can be close or far to the driver and the model needs to mitigate this randomness.
    \item \textbf{Horizontal Flip:} This doesn't correlate to real life setting, but the idea is that the driver's window will be always at its left side, which means the lightning conditions from the left side of the camera will be always poor compared to the right side. The model needs to detect fatigue regardless of the lightning hence the flip.
    \item \textbf{Image Fill:} Image can be scaled down or up. The model needs to detect fatigue regardless.
    \item \textbf{Scaling:} Image can be scaled down or up. The model needs to detect fatigue regardless.
\end{enumerate}

We develop two models, eye model and the face model where the first one is trained on eye patch dataset and the second one is trained on facial image dataset. For each model, we train a baseline model and a adversarially trained model.

\section{Related Work}

Recent years have been focusing on convolutional neural network based deep learning methods in order to detect eye closedness \cite{lk98} \cite{cg95} \cite{et98}. Before then, the focus was on identification of key points in face via edge detection \cite{c83}, based on 
these properties can be applied in face recognition \cite{a96} \cite{b92}. Among
these key points, the center of eyes is the most important than
others \cite{c83} \cite{mn98} \cite{lk98}, because with the use of eye centers one can remove
in-plane rotation of face and with calculating the distance
between eye centers, we can approximately find the size of
face in an image \cite{cg95} \cite{et98} \cite{cm91}. This is a helpful clue to find the location of
other key points \cite{ms98}. As a result, detection of eye centers
facilitates detection of other key points \cite{mn97} \cite{es99} \cite{es99measuringfacial}.
Some of the facial point extraction algorithms in literature
are: Template matching \cite{bp}, Integral projection \cite{ych92}, Snakes \cite{kwt87}, Deformable template \cite{bp}, Hough transform \cite{ych92}, Elastic
bunch graph matching \cite{ych92}, Region growing search \cite{ly96}, Active
Shape Models (ASMs), and Active Appearance Models
(AAMs) \cite{ly96}.

The recent years saw an shift on focus to convolutional neural network based deep learning methods. The focus on convolutional neural networks in order to detect the eye closedness has seen increase because of the improved accuracy as well as cheaper memory and faster CPU. In a convolutional neural network, each pixel is treated as a feature. A kernel is applied to an image that to perform a convolutional operation. Resulting matrix, then goes through max pooling with where each kernel-sized overlap produces one value, that is the maximum pixel value in that overlap.  Several kernels can be applied and the depth of the model can be very large like ResNet which has about 172 hidden layers. The goal is to extract features from a top-down setting where each layer produces the input for the following layer. After the Convolutional layers have been applied, the output is flattened and then given to a dense layer. The number of dense layers can be also arbitrary large, where each layer produces the input for the following layer. Both our models employ a 4 layer convolutional layer neural network. For a better understanding please refer to the Table \ref{tab:table3}

\section{Basic Approach}

We discussed that our goal is to reduce car accidents related to drowsy driving with an alert system. 

Neural network in the mobile application will alert the driver with sound when drivers are\\ 
i) closing their eyes frequently and long\\
ii) falling asleep on the wheel\\
iii) heads start to fall down\\
iv) drivers start to doze off\\
A robust model is needed to reduce the false positives. 
Model should not trigger when the driver might be \\
i) just blinking frequently \\
ii) leaning towards the radio\\
iii) checking mirrors/sideways \\

In order to achieve these goals, we build two convolutional neural network models, one trained on Eye-blink dataset, and one trained on CEW dataset. For easiness, we call the model trained on Eye-blink dataset, eye model because the Eye-blink dataset consists of only images of eye patches. For easiness, we call the model trained on the CEW dataset face model because the CEW dataset consists of facial images. We perform data augmentation and for both models.

\section{Data}

We use Eye-blink Dataset \cite{f14} \cite{pswl07} data set to  train our eye model. Eye Blink dataset consisting of 2100 closed and open eye images that are black and white. They are 24x24 pixels and only show eye patches. We use  Closed Eyes In The Wild (CEW) \cite{f14} dataset to train our face model. CEW dataset contains 1192 subjects with both eyes closed and 1231 subjects with eyes open. Some challenges of this set include amateur photography, occlusions, problematic lighting, pose, and motion blur.

\begin{figure}
\centering
  \includegraphics[width=0.9\columnwidth]{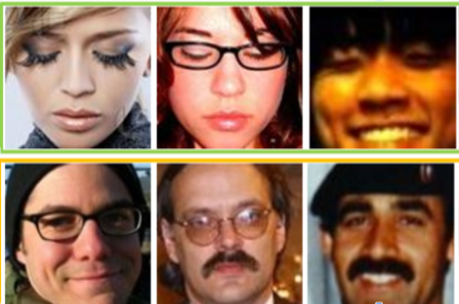}
  \caption{Samples from the Closed Eyes in the Wild Dataset (CEW), which contains 1192 subjects with both eyes closed and 1231 subjects with eyes open. Some challenges of this set include amateur photography, occlusions, problematic lighting, pose, and motion blur. Two models have been trained on this model: Face baseline model and adversarially trained model. 
 }~\label{fig:figure1}
\end{figure}

\begin{figure}
\centering
  \includegraphics[width=0.9\columnwidth]{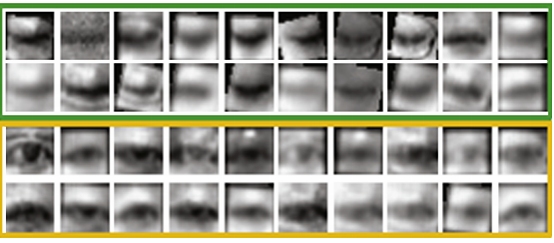}
  \caption{Samples from the Eye Blink dataset, which consists of 2100 closed and open 24x24 pixel grayscale eye patch images. Two models have been trained on this model: Eye baseline model and adversarially trained model.
 }~\label{fig:figure2}
\end{figure}

\section{Evaluation}
All four models have been trained and classification reports for the training and the testing datasets for both models can be seen in Table \ref{tab:table1} and Table \ref{tab:table2}.

\begin{table*}[ht] 
\caption{All four models (eye baseline, eye adversarial, face baseline, face adversarial) and their training results. Overall, the eye model has performed better than the face model and it can be seen in the comparison of the train and test accuracies of the eye baseline model to the train and test accuracies of the face baseline model.}
\centering
\scalebox{0.88}{
\begin{tabular}{|l|l|l|l|l|l|}
\hline
\textbf{Model Types} & \textbf{Test Accuracy} & \textbf{Test Precision}&\textbf{Test Recall}&\textbf{Test F1 Score}&\textbf{Test Loss}\\
\hline
Eye Baseline & 0.9584 & 0.9577 &0.9606 & 0.9575 & 0.1079 \\
Eye Adversarial & 0.9153 & 0.9184& 0.9190  & 0.9186 & 0.2224\\
\textbf{Face Baseline} & \textbf{0.9664} & \textbf{0.9682}&\textbf{0.9645} & \textbf{0.9646} & \textbf{0.0923}\\
Face Adversarial  & 0.8001 & 0.7945&0.8347 & 0.8042 & 0.4393\\
\hline
\end{tabular} \label{tab:table1}
}
\end{table*}

\begin{table*}[ht] 
\caption{All four models (eye baseline, eye adversarial, face baseline, face adversarial) and their testing results. Overall, the eye model has performed better than the face model and it can be seen in the comparison of the train and test accuracies of the eye baseline model to the train and test accuracies of the face baseline model. Also, adversarial face model performed the worst during testing.}
\centering
\scalebox{0.825}{
\begin{tabular}{|l|l|l|l|l|l|}
\hline
\textbf{Model Types} & \textbf{Train Accuracy} & \textbf{Train Precision}&\textbf{Train Recall}&\textbf{Train F1 Score}&\textbf{Train Loss}\\
\hline
\textbf{Eye Baseline} & \textbf{0.9476} & \textbf{0.9717} & 0.9097 &\textbf{0.9386} & \textbf{0.2358} \\
Eye Adversarial & 0.9126 & 0.8932&\textbf{0.9499}  & 0.9206& 0.4725\\
Face Baseline & 0.9271& 0.9583 &0.8863 & 0.9187 & 0.2561\\
Face Adversarial  & 0.7943  & 0.8197&0.7603 & 0.7831 & 0.4072 \\
\hline
\end{tabular} \label{tab:table2}
}
\end{table*}

\begin{table}[!htbp]
\caption{Model Architecture for both models. We used the same architecture on the Eye-Blink and CEW dataset. The eye model and the face model both use binary cross entropy with Adam's optimizer on iterative gradient descent.}
\label{tab:table3}
\centering
\scalebox{0.95}{
 \begin{tabular}{|c|c| c|} 
 \hline
 \textbf{Layer type} & \textbf{Output Shape} & \textbf{Param} \# \\ [0.5ex] 
 \hline
 Conv2D & (98, 98,6) & 60 \\ 
 \hline
 Average Pooling  & (49,49,6) & 0 \\
 \hline
 Conv2D & (47,47,16) & 880 \\
 \hline
 Average Pooling & (23,23,16) & 0  \\
 \hline
 Flatten & 8464 & 0 \\ 
 \hline
 Dense & 120 & 1015800 \\ 
 \hline
 Dense & 84 & 10164 \\ 
 \hline
 Dense & 1 & 85 \\ 
 \hline
\end{tabular}
}
\end{table}

\makeatletter
\let\origsection\section
\renewcommand\section{\@ifstar{\starsection}{\nostarsection}}

\newcommand\nostarsection[1]
{\sectionprelude\origsection{#1}\sectionpostlude}

\newcommand\starsection[1]
{\sectionprelude\origsection*{#1}\sectionpostlude}

\newcommand\sectionprelude{%
  \vspace{0em}
}

\newcommand\sectionpostlude{%
  \vspace{0em}
}
\makeatother
\subsection{Eye Model}
We developed a 4 layer convolutional neural network. We refer to this model as the "eye model". 

\subsection{Eye Model Results Approach 1: Noisified Training Data}
In our first approach, we added noise to the eye training dataset to simulate adversarial training under black-box settings as it can be seen in Table \ref{tab:eye_adversarial_training_parameters}

\begin{table}[!htbp]
\caption{Noisified training data parameters. The dataset is added noise to simulate adversarial settings. Adding noise to the training data simulates adversarial training under black-box settings without knowing the gradients of the model.}
\label{tab:eye_adversarial_training_parameters}
\centering
\scalebox{0.95}{
 \begin{tabular}{|c|c|} 
 \hline
 \textbf{Change Type} & \textbf{Change Amount} \# \\ [0.5ex] 
 \hline
 rotation range & 40$^o$\\ 
 \hline
  width shift range  & 0.2 \\
 \hline
 height shift range & 0.2 \\
 \hline
 shear range & 0.2\\
 \hline
 zoom range & 0.2\\ 
 \hline
 horizontal flip & True\\ 
 \hline
 fill mode & 'nearest' \\ 
 \hline
 rescale &  1./255\\ 
 \hline
\end{tabular}
}
\end{table}

\begin{figure}[!htbp]
\centering
  \includegraphics[width=1\columnwidth]{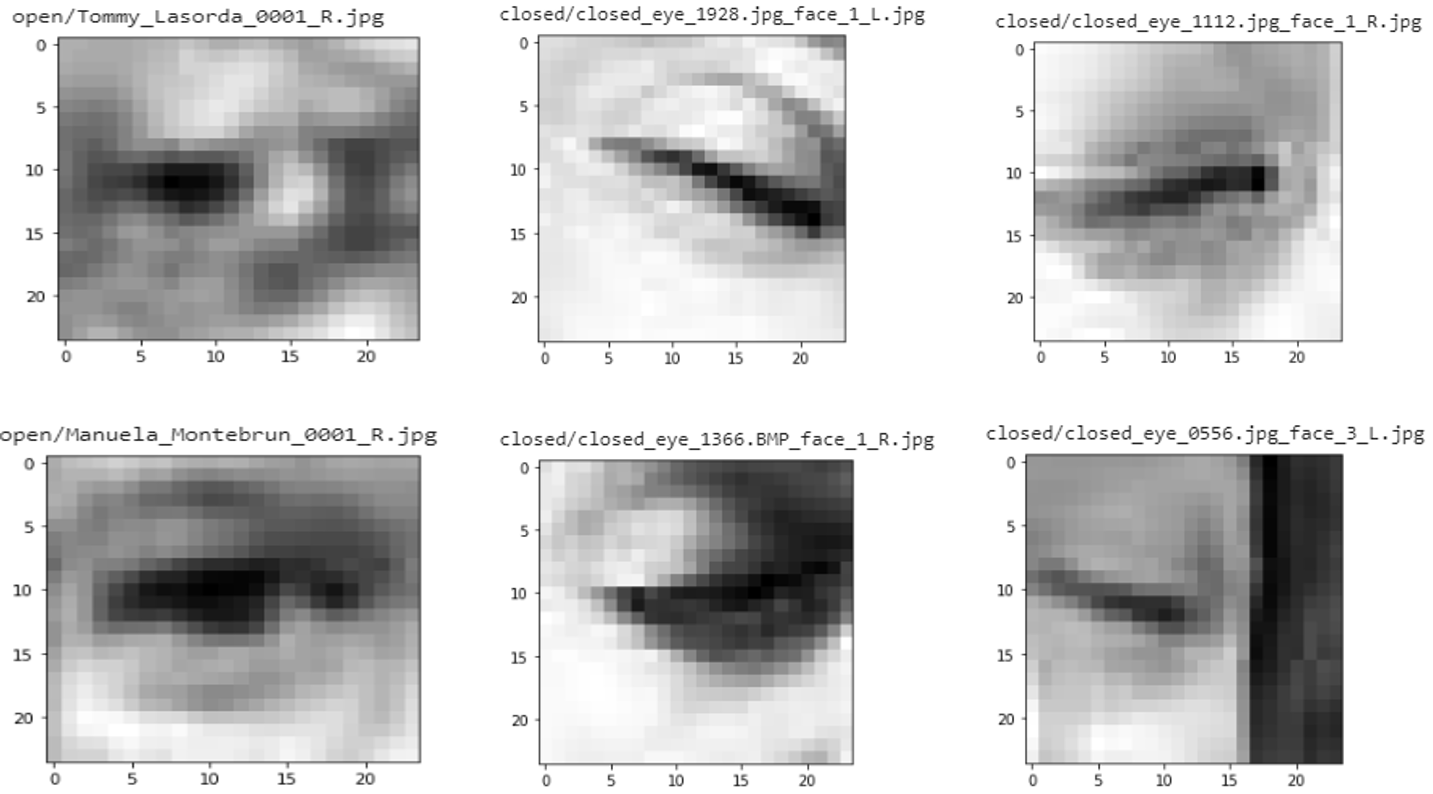}
  \caption{Noise added samples from the training dataset for the eye model. Adding noise emulate black-box attacks in real-life scenarios more closely and makes it easier for the models to generalize better to those scenarios. The perturbation added to the dataset simulates creating adversarial attacks. After adding adversarial perturbations and noise to the training dataset, the dataset is used to train a model and increase robustness.}
  \label{fig:figure4}
\end{figure}

\begin{figure}[!htbp]%
    \centering
    \subfloat[Eye model accuracy for training and testing dataset versus epoch count when \textbf{noise is added.} ]{\includegraphics[scale=0.70]{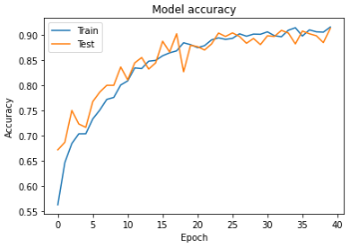} }
    \qquad
    \subfloat[Eye model binary cross entropy loss for training and testing datasets versus epoch count when \textbf{noise is added.}]{\includegraphics[scale=0.70]{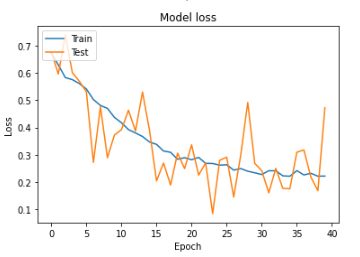} }%
    \caption{Training and testing of eye model with Eye-blink dataset. Due to the existence of adversarially added noise in the dataset, this model is referred as adversarially trained eye model.}
    \label{fig:accuracy_loss_adversarial_eye}
\end{figure}



\begin{table}[!htbp]
\caption{Eye model accuracy, f1 score, recall and precision when \emph{noise is added} to the Eye-blink dataset. Class label 0 is for when the eyes are closed. Class label 1 is for when the eyes are open.}
\centering
\scalebox{0.93}{
 \begin{tabular}{|c|c|c|c|c|c|} 
 \hline
 \textbf{Class Label} & \textbf{Accuracy} & \textbf{Precision} & \textbf{Recall} & \textbf{F1 Score} & \textbf{Sample \#} \\ [0.5ex] 
 \hline
 \textbf{0} & 0.93 & 0.95 & 0.90 & 0.93 & 470 \\ 
 \hline
 \textbf{1} & 0.93 & 0.91 & 0.96 & 0.93 & 492 \\ 
 \hline
\end{tabular}
}
\end{table}

\begin{table}[!htbp]
\caption{Confusion matrix for the eye model when \emph{noise is added} to the Eye-blink dataset. Class label 0 is for when the eyes are closed. Class label 1 is for when the eyes are open.}
\centering
\scalebox{1.3}{
 \begin{tabular}{|c|c|c|} 
 \hline

 \diagbox{\textbf{True}}{\textbf{Predicted}}& \textbf{Class 0} & \textbf{Class 1}  \\ 
 \hline
 \textbf{Class 0} & 424 & 46 \\
 \hline
 \textbf{Class 1} & 21 & 471 \\ 
 \hline
\end{tabular}
}
\vspace{-5mm}
\end{table}

We trained for 40 epochs with 3108 images belonging to 2 classes, eye closed and open (same number on both classes). We then tested the model on 776 images.

\newpage
\begin{enumerate}
    \item Testing data for the adversarial eye model had higher accuracy than training data of the same adversarial eye model. 
    \item Potential cause for underfitting.
    \item Training data had several arduous cases to learn due to noise
    \item Test data contained easier cases to predict
    \item Training loss is the average of the losses over each batch of training data. 
\end{enumerate}
Because the model gradients are changing over time, the loss over the first batches of an epoch is generally higher than over the last batches. On the other hand, the testing loss for an epoch is computed using the model as it is at the end of the epoch, resulting in a lower loss.

\subsection{Eye Model Results Approach 2: No noise addition}
 
We trained for 40 epochs with 3108 images belonging to 2 classes, eye closed and open (same number on both classes). Later, the model is tested on 776 images.

\begin{enumerate}
    \item Training data had higher accuracy than training data.(see red highlights)
    \item This might be due to the fact that the model is overfitting.
    \item Removing data augmentation in this run yielded 6\% improvement in accuracy, precision, recall and the f1-score.
\end{enumerate}

\begin{figure}[!htbp]%
    \centering
    \subfloat[Eye model accuracy for training and testing dataset versus epoch count when \emph{no noise is added.} This model is referred as the baseline eye model. ]{\includegraphics[scale=0.70]{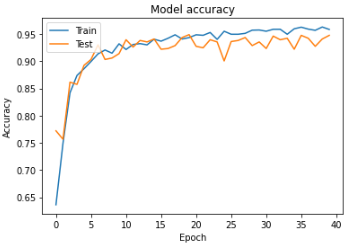} }
    \qquad
    \subfloat[Eye model binary cross entropy loss for training and testing datasets versus epoch count when \emph{no noise is added.}]{\includegraphics[scale=0.70]{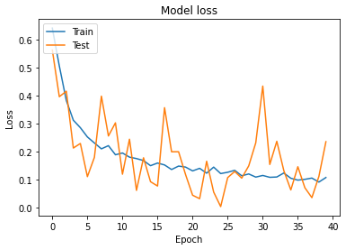} }%
    \caption{Training and testing of baseline eye model with Eye-blink dataset. This model is referred as baseline eye model.}
    \label{fig:accuracy_loss_baseline_eye}
\end{figure}

\begin{table}[!htbp]
\caption{Eye model accuracy, f1 score, recall and precision when \emph{no noise is added} to the Eye-blink dataset. Class label 0 is for when the eyes are closed. Class label 1 is for when the eyes are open.}
\centering
\scalebox{0.93}{
 \begin{tabular}{|c|c|c|c|c|c|} 
 \hline
 \textbf{Class Label} & \textbf{Accuracy} & \textbf{Precision} & \textbf{Recall} & \textbf{F1 Score} & \textbf{Sample \#} \\ [0.5ex] 
 \hline
 \textbf{0} & 0.95 & 0.93 & 0.97 & 0.95 & 470 \\ 
 \hline
 \textbf{1} & 0.95 & 0.97 & 0.93 & 0.95 & 492 \\ 
 \hline
\end{tabular}
}
\end{table}

\begin{table}[!htbp]
\caption{Confusion matrix for the eye model when \emph{no noise is added} to the Eye-blink dataset. Class label 0 is for when the eyes are closed. Class label 1 is for when the eyes are open.}
\centering
\scalebox{1.3}{
 \begin{tabular}{|c|c|c|} 
 \hline

 \diagbox{\textbf{True}}{\textbf{Predicted}}& \textbf{Class 0} & \textbf{Class 1}  \\ [0.5ex] 
 \hline
 \textbf{Class 0} & 455 & 15 \\
 \hline
 \textbf{Class 1} & 33 & 459 \\ 
 \hline
\end{tabular}
}
\vspace{-5mm}
\end{table}

\subsection{Face Model}
We developed a 4 layer convolutional neural network. We refer to this model as the "face model".

 \subsection{Face Model Results Approach 1: Noisified Training Data}
 In our first approach, we added noise to the face training dataset to simulate adversarial training under black-box settings as it can be seen in Table \ref{tab:face_adversarial_training_parameters}
 
\begin{table}[!htbp]
\caption{Noisified training data parameters are shown below. Adding noise to the training data simulates adversarial training under black-box settings without knowing the gradients of the model.}
\label{tab:face_adversarial_training_parameters}
\centering
\scalebox{0.95}{
 \begin{tabular}{|c|c|} 
 \hline
 \textbf{Change Type} & \textbf{Change Amount} \# \\ [0.5ex] 
 \hline
 rotation range & 40$^o$\\ 
 \hline
  width shift range  & 0.2 \\
 \hline
 height shift range & 0.2 \\
 \hline
 shear range & 0.2\\
 \hline
 zoom range & 0.2\\ 
 \hline
 horizontal flip & True\\ 
 \hline
 fill mode & 'nearest' \\ 
 \hline
 rescale &  1./255\\ 
 \hline
\end{tabular}
}
\vspace{1em}
\end{table}

\begin{figure}[!htbp]
\centering
  \includegraphics[width=1\columnwidth]{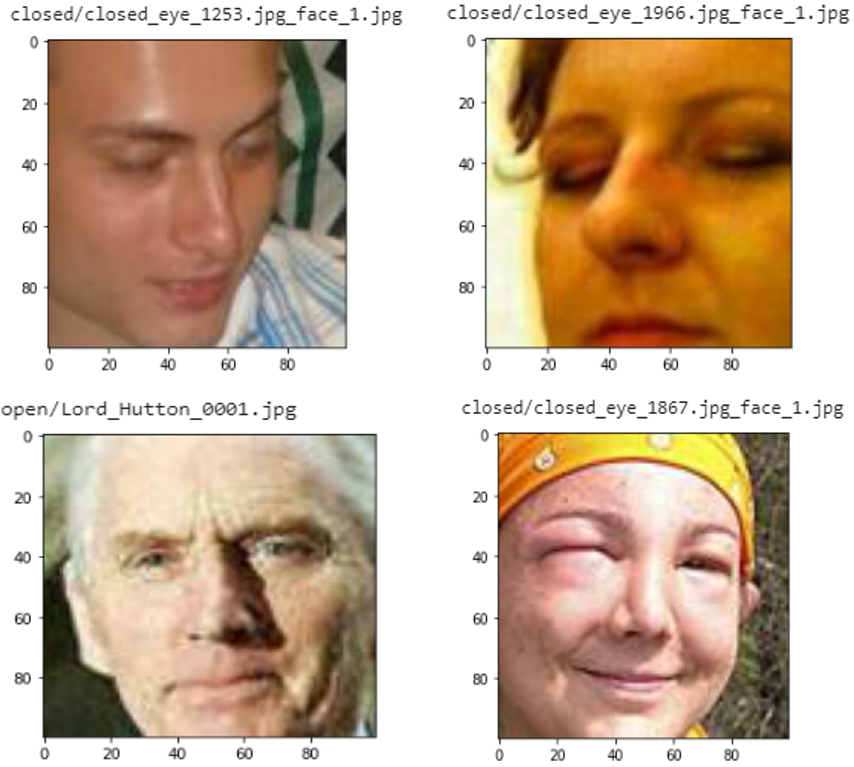}
  \caption{Noise added samples from the training dataset used for the face model. Adding noise emulate black-box attacks in real-life scenarios more closely and makes it easier for the models to generalize better to those scenarios.   The perturbation added to the dataset simulates creating  adversarial  attacks.   After  adding  adversarial  perturbations  and  noise  to  the training dataset, the dataset is used to train a model and increase robustness }~\label{fig:figure12}
\end{figure}

\begin{figure}[!htbp]%
    \centering
    \subfloat[Face Model Accuracy for training and testing dataset versus epoch count when \emph{noise is added.} This model is referred as the adversarially trained face model. ]{\includegraphics[scale=0.70]{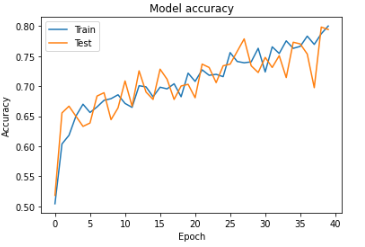} }
    \qquad
    \subfloat[Adversarially trained face model binary cross entropy loss for training and testing datasets versus epoch count when \emph{noise is added}.]{\includegraphics[scale=0.70]{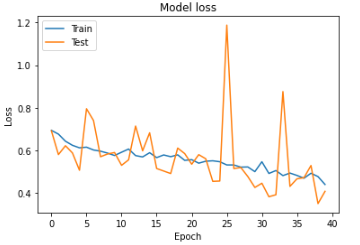} }%
    \caption{Training and testing of face model with CEW dataset. Due to the existence of adversarially added noise in the dataset, this model is referred as adversarially trained face model.}
    \label{fig:accuracy_loss_adversarial_face}
\end{figure}

\begin{table}[!htbp]
\caption{Face model accuracy, f1 score, recall and precision when \emph{noise is added} to the CEW dataset. Class label 0 is for when the eyes are closed. Class label 1 is for when the eyes are open.}
\centering
\scalebox{0.93}{
 \begin{tabular}{|c|c|c|c|c|c|} 
 \hline
 \textbf{Class Label} & \textbf{Accuracy} & \textbf{Precision} & \textbf{Recall} & \textbf{F1 Score} & \textbf{Sample \#} \\ [0.5ex] 
 \hline
 \textbf{0} & 0.93 & 0.92 & 0.93 & 0.93 & 239 \\ 
 \hline
 \textbf{1} & 0.93 & 0.93 & 0.92 & 0.93 & 246 \\ 
 \hline
\end{tabular}
}

\end{table}

\begin{table}[!htbp]
\caption{Confusion Matrix for the Face Model when \emph{noise is added} to the CEW dataset. Class label 0 is for when the eyes are closed. Class label 1 is for when the eyes are open.}
\centering
\scalebox{1.3}{
 \begin{tabular}{|c|c|c|} 
 \hline

 \diagbox{\textbf{True}}{\textbf{Predicted}}& \textbf{Class 0} & \textbf{Class 1}  \\ [0.5ex] 
 \hline
 \textbf{Class 0} & 223 & 16 \\
 \hline
 \textbf{Class 1} & 19 & 227 \\ 
 \hline
\end{tabular}
}
\end{table}

Overall, adversarial model performs poorly compared to baseline model. Especially, training accuracy for the adversarial face model is approximately 15\% lower than training accuracy of the baseline face model.
We trained the face model for 40 epochs to be consistent with the eye model over 1558 images belonging to 2 classes, eye closed or open. We then tested the model on 485 images.

\begin{enumerate}
    \item Testing data had higher accuracy than training data.(see red highlights)
    \item This might be due to the fact that the model is underfitting.
    \item The training data had several arduous cases to learn due to noise
    \item The test data contained easier cases to predict.
    \item high accuracy can be attributed to the fact that not memorizing a complex training data is useful for real life models.

\end{enumerate}

\subsection{Face Model Results Approach 2: No noise addition}

We trained the model for 40 epochs to be consistent with the eye model over 1558 images belonging to 2 classes, eye closed or open. We then tested the model on 485 images.

\begin{enumerate}
    \item Better results compared to noisy training data.
    \item Training accuracy is higher than testing accuracy.
    \item Training data had higher accuracy than training data.(see red highlights)
    \item The baseline model observes overfitting.
    \item Removing data augmentation in this run yielded 12\% improvement in accuracy, precision, recall and the f1-score.
\end{enumerate}

\begin{figure}[!htbp]%
    \centering
    \subfloat[Baseline face model accuracy for training dataset versus epoch count when \emph{no noise is added.} This model is referred as baseline face model because there is no adversarial attacks in the training dataset. ]{\includegraphics[scale=0.70]{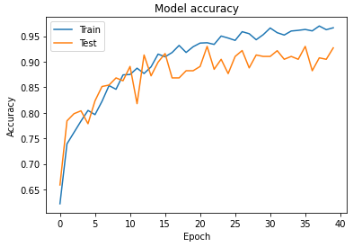} }
    \qquad
    \subfloat[Baseline face model binary cross entropy loss for testing datasets versus epoch count when \emph{no noise is added.}]{\includegraphics[scale=0.70]{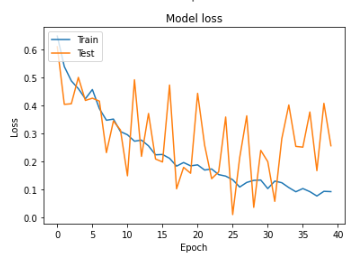} }%
    \caption{Training and testing of baseline face model with CEW dataset. This model is referred as baseline face model.}
    \label{fig:accuracy_loss_baseline_face}
\end{figure}

\begin{table}[!htbp]
\caption{Face model accuracy, f1 score, recall and precision when \emph{no noise} is added to the CEW dataset. Class label 0 is for when the eyes are closed. Class label 1 is for when the eyes are open.}
\centering
\scalebox{0.93}{
 \begin{tabular}{|c|c|c|c|c|c|} 
 \hline
 \textbf{Class Label} & \textbf{Accuracy} & \textbf{Precision} & \textbf{Recall} & \textbf{F1 Score} & \textbf{Sample \#} \\ [0.5ex] 
 \hline
 \textbf{0} & 0.93 & 0.92 & 0.93 & 0.93 & 239 \\ 
 \hline
 \textbf{1} & 0.93 & 0.93 & 0.92 & 0.93 & 246 \\ 
 \hline
\end{tabular}
}
\end{table}

\begin{table}[!htbp]
\caption{Confusion matrix for the face model when \emph{no noise} is added to the CEW dataset. Class label 0 is for when the eyes are closed. Class label 1 is for when the eyes are open.}
\centering
\scalebox{1.3}{
 \begin{tabular}{|c|c|c|} 
 \hline

 \diagbox{\textbf{True}}{\textbf{Predicted}}& \textbf{Class 0} & \textbf{Class 1}  \\ [0.5ex] 
 \hline
 \textbf{Class 0} & 223 & 16 \\
 \hline
 \textbf{Class 1} & 19 & 227 \\ 
 \hline
\end{tabular}
}
\end{table}

\section{Conclusion}
In this study, we explored two different approaches for driver fatigue detection. First, we trained a neural network on eye patch images to detect eyes closedness. Second, we trained a neural network on facial images to detect eyes closedness. This was a more relaxed version of the facial fatigue detection problem. The eye model performed better than the face model by achieving  94\% testing accuracy. This testing accuracy was 2\% higher compared to that of the face model. We also studied facial fatigue detection under an adversarial detection by applying data augmentation in order to increase robustness. Some of the perturbations that we added to the dataset for data augmentation were shear angle shift, zoom, rescaling. We considered these perturbations as black-box adversarial attacks and performed adversarial training to emulate real life driving settings. Adversarially trained models performed 4\% and 16\% worse on the training data and 3\% and 13\% worse on the testing data. This is due to the fact that we introduced arduous cases to the training images and made it harder to memorize the training data hence poor training accuracies. But they generalized well so we were able to see less performance decreases in the testing accuracies compared to the testing accuracies of the non-adversarial models. One important thing to notice is that we were able to increase recall for the eye closedness detection by about 5\%. Recall, which quantifies the number of positive class predictions made out of all positive examples in the dataset suggests that if the task is small enough, introducing adversarial training can reduce the false negatives and increase the true positives, which is very crucial for drowsy driver detection because not being able to detect fatigue when it is most needed is very crucial to prevent car accidents related to drowsiness. We were also able to freeze the weights in our all models and convert them to \textit{tflite} files that can be used in a mobile application. When we converted the models to \textit{tflite} models \cite{tffreemodels}, the model size was approximately ~19 MB, which is a size that can be put into a mobile application. Since we used the same architecture for all four models, we achieved the same file size for all four models. We hope that these remarks will help future researchers.

\section{Future Work}
In the future,  we would like to explore ensemble learning and combine the two models that we had. We can produce a hybrid model where the eyes closedness is classified by votes coming from the eye model as well as the face model. We also would like to explore the relative head position in order to better detect the driver's fatigue instead of just focusing on the eye closedness detection.
We also think it would be interesting to train an eye detector by using the face dataset, then use the eye dataset eyes to train an eye closedness detector and compare that to just the eye detector. This comparison would highlight the differences between using only eye images and eye + face images for eye detection.

\section{Acknowledgments}

We offer our most sincere appreciation to Dr. Matthew Burlick, who gave valuable feedback during several phases of our research in this study.

\printbibliography[title=List of References]

\end{document}